%% file: alert_paper.tex
\pdfoutput=1

\documentclass{article}

\usepackage{microtype}
\usepackage{graphicx}
\usepackage{subfigure}
\usepackage{booktabs} 
\usepackage{tablefootnote}

\usepackage{hyperref}

\usepackage[accepted]{icml2024/icml2024}

\usepackage{amsmath}
\usepackage{amssymb}
\usepackage{mathtools}
\usepackage{amsthm}

\usepackage{pifont}
\newcommand{\cmark}{\ding{51}}%
\newcommand{\xmark}{\ding{55}}%

\usepackage{multirow}

\usepackage[capitalize,noabbrev]{cleveref}

\theoremstyle{plain}

\theoremstyle{definition}

\theoremstyle{remark}

\usepackage[disable,textsize=tiny]{todonotes}

\icmltitlerunning{ALERT-Transformer}

\begin{document}

\twocolumn[
    \icmltitle{
        ALERT-Transformer: Bridging Asynchronous and Synchronous Machine Learning for Real-Time Event-based Spatio-Temporal Data
    }



    \icmlsetsymbol{equal}{*}

    \begin{icmlauthorlist}
        \icmlauthor{Carmen Martin-Turrero}{equal,sony,unip}
        \icmlauthor{Maxence Bouvier}{equal,sony}
        \icmlauthor{Manuel Breitenstein}{sony}
        \icmlauthor{Pietro Zanuttigh}{unip}
        \icmlauthor{Vincent Parret}{sony}
    \end{icmlauthorlist}

    \icmlaffiliation{sony}{Sony Semiconductor Solutions Europe, Sony Europe B.V, Stuttgart Laboratory 1, Zurich, Switzerland}
    \icmlaffiliation{unip}{University of Padova, MEDIA Lab, Veneto, Italy}

    \icmlcorrespondingauthor{Vincent Parret}{vincent.parret@sony.com}

    \icmlkeywords{Machine Learning, Computer Vision, Event-Based Data, Transformer Networks, Point Cloud Networks, ICML}

    \vskip 0.3in
]



\printAffiliationsAndNotice{\icmlEqualContribution} 

\begin{abstract}
    We seek to enable classic processing of continuous ultra-sparse spatiotemporal data generated by event-based sensors with dense machine learning models.
    We propose a novel hybrid pipeline composed of asynchronous sensing and synchronous processing that combines several ideas: (1) an embedding based on PointNet models -- the ALERT module -- that can continuously integrate new and dismiss old events thanks to a leakage mechanism, (2) a flexible readout of the embedded data that allows to feed any downstream model with always up-to-date features at any sampling rate, (3) exploiting the input sparsity in a patch-based approach inspired by Vision Transformer to optimize the efficiency of the method.
    These embeddings are then processed by a transformer model trained for object and gesture recognition.
    Using this approach, we achieve performances at the state-of-the-art with a lower latency than competitors.
    We also demonstrate that our asynchronous model can operate at any desired sampling rate.
\end{abstract}

\input{sections/introduction}

\input{sections/relatedwork}

\input{sections/methodology}

\input{sections/results}

\input{sections/discussion}

\input{sections/conclusions}






\section*{Impact Statement}

This paper presents work whose goal is to advance the field of Machine Learning. There are many potential societal consequences of our work. We shortly mention two of them here specifically related to using an Event-based technology. Our work aims at bridging the gap between asynchronous, event-based signal processing and conventional signal processing. On one hand, in doing so, it facilitates the use of such low-latency system, for example for autonomous systems, and make them more robust and possibly safe (e.g. for obstacle detection). On the other hand, accelerating the use of event-based sensing and processing in computer vision might lead to systems requiring less power, which, at scale, is relevant for the environment. We do not foresee any negative impact of this work.

\bibliography{alert_paper}
\bibliographystyle{icml2024/icml2024}

\newpage
\appendix
\onecolumn
\input{sections/appendix}

\end{document}

%% file: sections/introduction.tex
%
%

\section{Introduction}
\label{sec:introduction}

Event-based sensors capture visual information in an event-driven, asynchronous manner \cite{finateu_510_2020, gallego_event-based_2022}.
Efficiently exploiting their data has proven challenging as the vast majority of approaches published in the literature consist of either converting event-based data to dense representations, or deploying spiking neural networks (SNNs) on streams of events.
The former allows to exploit standard machine learning (ML) frameworks such as \texttt{PyTorch} and \texttt{Tensorflow}, but does not leverage the inherent sparsity and other properties of event-based data \cite{gehrig_end--end_2019}.
The latter relies on SNNs, which are hard to train and usually exhibit lower accuracy than an equivalent dense neural network.
Furthermore, while the neuromorphic community has argued in favor of their higher energy efficiency for decades, recent research and breakthroughs in edge AI accelerators indicate this is still an open question \cite{dampfhoffer_are_2023, garrett_1mw_2023, moosmann_ultra-efficient_2023, caccavella_low-power_2023}.

Nevertheless, considering the inherent advantages of event-based vision sensors, namely high dynamic range (HDR) and high temporal resolution -- simultaneously, without any tradeoffs between the two --, we aim to find a way to leverage this sparse and low-latency data for real-world situations.

\begin{figure*}[h]
	\centering
	\includegraphics[width=\textwidth]{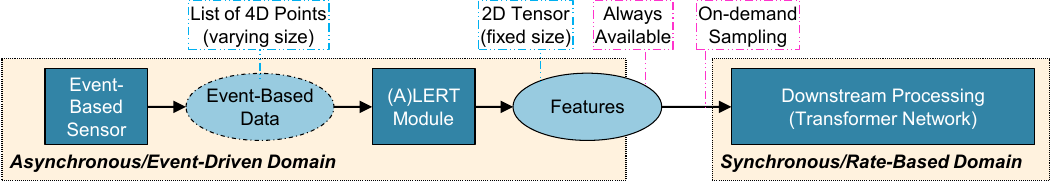}
	\vskip -0.1in
	\caption{Schematic representation of the system integrating our proposed asynchronous embedding module. The asynchronous part (left) processes all events as they come in an event driven manner, thus updating the $Features$ tensor continuously. The synchronous part (right) samples the $Features$ tensor on demand, thus allowing a seamless interface between asynchronous and synchronous processing.}
	\label{fig:async_to_syn}
	\vskip -0.1in
\end{figure*}

Standard ML relies on tensor-based processing.
Converting the stream of events -- represented as tuples of values ($x$ and $y$ pixel coordinates, polarities and timestamp) -- to a multidimensional tensor is thus a crucial step.
The challenge involves (1) representing time in a reliable and continuous manner, allowing it to be processed similarly to the finite spatial and polarity dimensions, (2) continuously incorporating new events in the feature tensors which also requires forgetting previous events, (3) using limited computational resources to allow real-time processing.
Our main contributions towards Event-Based ML are the following:

\begin{itemize}
	\itemsep-0.05in
	\item The ALERT module, an embedding based on PointNet which continuously integrates new events dismissing old ones via a leakage mechanism. This module introduces novel asynchronous embedding updates.
	\item A flexible readout of the embedded data that can feed any downstream model with up-to-date features at different sampling rates, down to a per-event operation, allowing ultra-low latency decision making.
	\item A patch-based approach inspired by Vision Transformer to exploit input sparsity and optimize efficiency.
	\item A time encoding solution to represent continuous time as a bi-dimensional vector of bounded values, at the cost of negligible decrease of relative accuracy.
	\item The ALERT-Transformer, a framework incorporating all of the above, which is trained on event-based data end-to-end. The model can then operate in synchronous regime for high accuracy on gesture recognition, or asynchronously for ultra-low latency.
	\end{itemize}

Our idea is that, even if the sensor outputs data at 1MHz or more \cite{finateu_510_2020}, the application/client side will not, and does not need to, run at this speed.
For instance, in an embedded product, such as a drone, that relies on event-based sensors for self-localization in space, this localization step would not operate at more than 100Hz \cite{kaufmann_champion-level_2023}.
Therefore, we consider that this hybrid (asynchronous-to-synchronous) conversion module would be operated as close to the sensor as possible, inside a 3D-stacked image sensor, as proposed in \cite{bouvier_scalable_2021, bonazzi_tinytracker_2023}.

%% file: sections/relatedwork.tex
%
%

\subsection{Related Works}

\subsubsection{Event-Based Data Representations}
\label{sec:EVS}

Event-based data is extremely sparse, with unpredictable sparsity patterns, making it unsuitable for raw processing with dense machine learning pipelines.
An extended approach consists in integrating events over fixed time windows, creating frame-like representations using methods such as histograms or event queues \cite{innocenti_CNNframesGesture, Sabater_2022_CVPR, Maqueda_2018, evflownet}.
To maximize task accuracy from end-to-end, \cite{gehrig_end--end_2019} proposed learning the kernel for convolving the stream of events into a discrete tensor, called ``Event-Spike Tensor".
While the resulting networks perform better, this method still requires waiting to accumulate all events before starting processing and is thus not asynchronous.

\subsubsection{PointNet Architecture}
The PointNet \cite{qi2016pointnet, qi_pointnet_2017} is an architecture designed for processing three-dimensional point clouds.
It extracts features from individual points with a shared Multi-Layer Perceptron (MLP).
It then reduces the entire point cloud with a max-pooling operation into a single global feature vector.
Given the similarity in data representation, several works interpreted event streams as point clouds \cite{wang_clouds, humanpose_pointtransformer, zhao2021point, dgcnn}.
These approaches mostly treat time as any other coordinate, making fast and efficient processing challenging.

EventNet \cite{sekikawa_eventnet_2019} addresses this issue by processing time separately from other coordinates.
They propose a modification to the $max$ operator which becomes a recursive function capable of updating temporal and spatial information as new events arrive.
Our solution, introduced in Section \ref{sec:methodology}, eliminates the need to modify the $max$ operator.
The structure of the PointNet remains unaltered and is thus entirely compatible with existing AI accelerators.

\subsubsection{Event-Based Transformers}
Our work takes inspiration from PointBERT \cite{yu_point-bert_2022}, which combines PointNets with a Transformer \cite{DBLP:journals/corr/VaswaniSPUJGKP17} for 3D classification.
Models proposed by \cite{Sabater_2022_CVPR, sabater2022event, wang_exploiting, GET_peng} generate frames of aggregated events using various strategies, and convert them into tokens with a patch-to-token strategy inspired by the Vision Transformer (ViT\footnote{In ViT, images are split into fixed-size patches, which are then linearly embedded to obtain a sequence of vectors.}) \cite{vit}.
They leverage spatial sparsity by discarding input patches lacking sufficient events.
In \cite{GET_peng}, grouped convolutions are used to embed successive patches through time, each token thus handles information from different time steps.
\cite{humanpose_pointtransformer} attempts to input a reduced 3D event cloud to the Point Transformer \cite{zhao2021point}, a model originally designed for spatial 3D point clouds.
Following a different approach, \cite{Blegiers2023EventTransAct} employ a Video Transformer Network using event-frames.
All these approaches demonstrate high accuracy, motivating our choice of using Transformer models.
However, most are not trained end-to-end, and none can process events asynchronously.

\subsubsection{Asynchronous Processing}
Besides \cite{sekikawa_eventnet_2019}, some works focus on asynchronous processing of event streams with standard network architectures.
\cite{HATS_sironi} presented an architecture using local memory units shared by neighboring pixels.
\cite{messikommer} introduced Asynchronous Sparse Convolutional Networks, a framework for converting models trained on synchronous image-like event representations into asynchronous models.
Even though these solutions are built for event-driven processing, they add on complexity with respect to standard ML tools.

%% file: sections/methodology.tex
%
%

\section{Proposed Architecture}
\label{sec:methodology}
Our objective is to employ ML techniques to build input embeddings from an asynchronous stream of events, while preserving the properties of event-based data.
For this, we combine a PointBERT architecture \cite{yu_point-bert_2022} with an embedding inspired by EventNet \cite{sekikawa_eventnet_2019}.

\begin{figure}[h]
    \begin{center}
        \centerline{\includegraphics[width=\columnwidth]{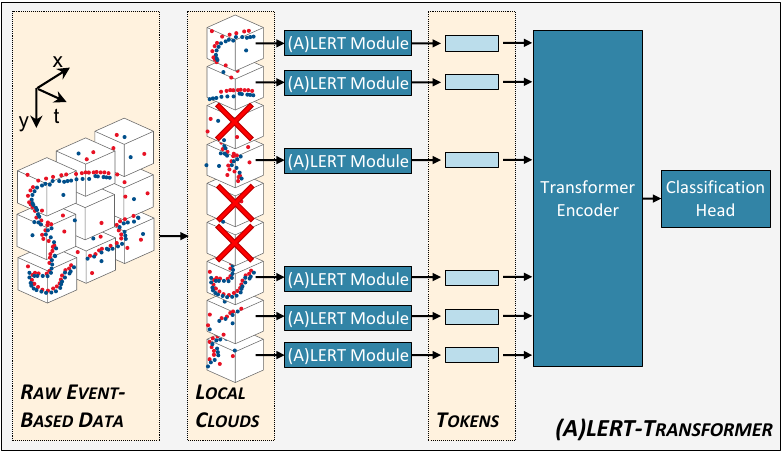}}
        \caption{Overview of the (A)LERT-Transformer model. The event stream is spatially divided into local event clouds. The red crosses indicate the non-active tokens based on their number of events. (A)LERT module converts them to individual high dimension features, which are fed to a Transformer and classifier head.}
        \label{fig:LERT-Transformer}
    \end{center}
    \vskip -0.2in
\end{figure}

\subsection{LERT: Synchronous Events-To-Tokens Conversion}
\subsubsection{Raw Events to Normalized Local Patches}
The embedding part -- which we call \textbf{L}earnt \textbf{E}mbedding for \textbf{R}eal-\textbf{T}ime Processing of Event-Based Data (LERT) Module -- is built upon the PointNet architecture \cite{qi2016pointnet}.
LERT spatially divides the input stream in local event clouds (patches) and converts each of them into individual feature vectors (a.k.a.\ tokens). This module is depicted in Figure \ref{fig:embedding_strategy}.
To extract patches from an event sequence, a grid in the (x, y)-plane is constructed, effectively dividing the space into same-sized groups of pixels with predefined coordinates. Events triggered in the same pixel-group make a patch. Hence, a local event cloud is simply a set of events, which is a subset of the original input event stream.

\begin{figure}[h]
    \begin{center}
        \centerline{\includegraphics[width=\columnwidth]{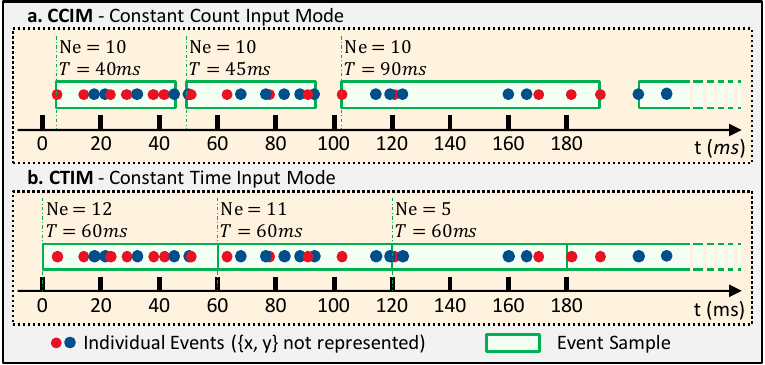}}
        \caption{Schematic representation of the temporal sampling. In (a) Constant Count Input Mode (CCIM), the events are sampled from the continuous input event stream by splitting them into bins that include the same amount of events. In (b) Constant Time Input Mode (CTIM), each bin represents the same input duration. In both modes, every sample is then split into several subsets, based on the spatial coordinates of the events, as depicted in Figure \ref{fig:LERT-Transformer}. Note: blue and red points represent events with different polarities.}
        \label{fig:input_modes}
    \end{center}
    \vskip -0.2in
\end{figure}

Once the events are organized in patches, the LERT module applies two transformations.
First, all patches which do not contain enough events are filtered out.
The filter mechanism is a simple threshold on the number of events present inside the local point cloud. The threshold value is fixed and selected via a hyperparameter search. It could be trained.
We denote the removed patches and events they contain as \emph{non-active}.
This method was proposed by \cite{Sabater_2022_CVPR} and allows to exploit the sparsity of the input event-based data.
The remaining events and patches are referred to as \emph{active} events and active patches.
The second step is a trivial normalization, where the spatial coordinates $(x, y)$ of the active events are scaled down to the range $[-1, 1]$ with respect to a patch size (\emph{not} the full image size).
Once this preprocessing of the events is finished, the active event coordinates inside each patch are represented as follows:
\begin{equation}
    \begin{split}
        (t, x, y, p) \mapsto & \ t\in[0,\ T],\ x\in[-1,\ 1], \\
        & \ y\in[-1,\ 1],\ p\in\{-1,\ 1\};
    \end{split}
\end{equation}
$T$ denotes the duration of the input event stream sample (\emph{not} the full file duration).
LERT operates in a fully synchronous manner which makes it compatible with widespread machine learning tools and frameworks, simplifying training.
It takes as input a finite list of events whose length ($Ne$) depends on the event accumulation mode.
$Ne$ is either always the same (we call it \emph{Constant Count Input Mode} (\textbf{CCIM})), or varying and depending on the number of events triggered during a fixed time window (denoted \emph{Constant Time Input Mode} (\textbf{CTIM})), in which case $T$ is always the same between two samples. Figure \ref{fig:input_modes} provides a visualization of event data and the two possible input modes.

\subsubsection{Normalized Local Patches to Tokens}
The Feature Generator (FG) then converts each normalized input patch into an input token for the transformer model.
The FG is shared between all the patches and consists in a customized shallow PointNet architecture.
Every 4D active event contained in a single patch is processed through a shared MLP unit to generate a single high dimensional vector (of dimension $c$).
The feature generator inherits the capacity of the PointNet to process an input of varying length.
In our implementation, an MLP is composed of a succession of 1D-convolution layers.

\begin{figure*}[t]
    \centering
    \includegraphics[width=\textwidth]{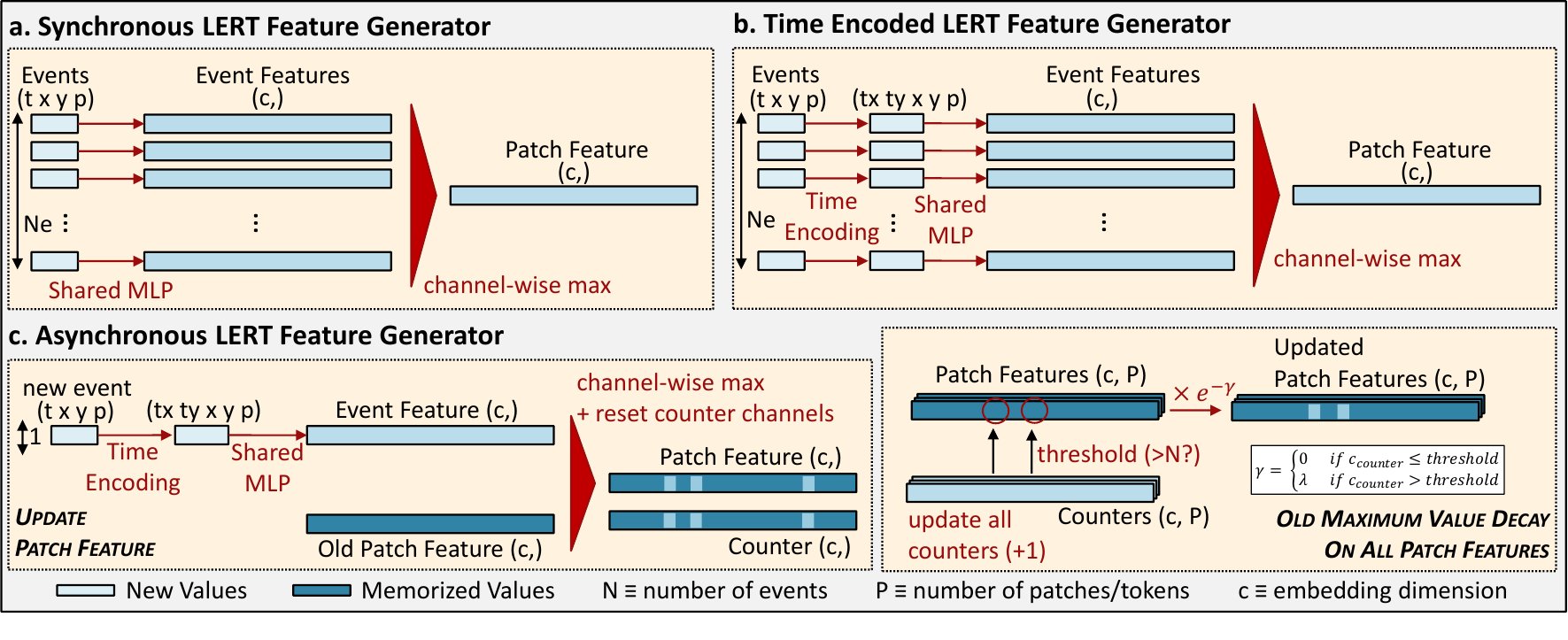}
    \vskip -0.2in
    \caption{Overview of the modes of the (A)LERT feature generator (FG). The Time Encoded LERT FG (b) is used for training ALERT (c).}
    \label{fig:featuregen}
    \vskip -0.1in
\end{figure*}

Each 4D event is thus mapped to a high dimensional vector, and the Patch Feature is obtained by applying a channel-wise maximum operation on the resulting vectors.
At this stage, all $P$ active patches have been converted into $P$ $c$-dimensional vectors, denoted as Patch Features (Fig. \ref{fig:featuregen} (a)).

\begin{figure}[h]
    \begin{center}
        \centerline{\includegraphics[width=\columnwidth]{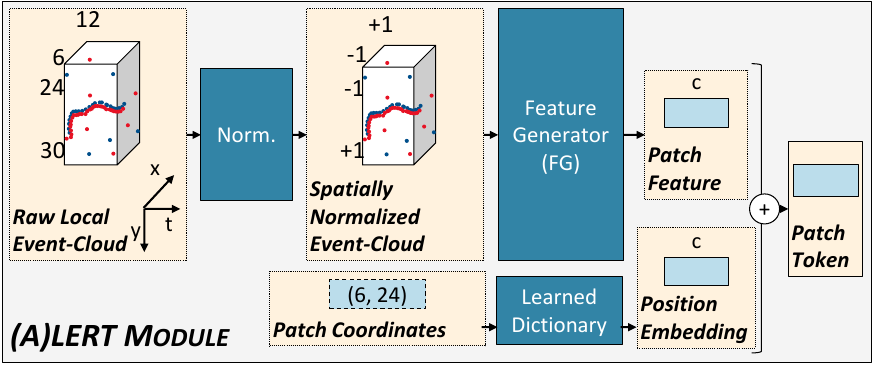}}
        \caption{(A)LERT module: spatially local event cloud to token.}
        \label{fig:embedding_strategy}
    \end{center}
    \vskip -0.2in
\end{figure}

Normalizing and scaling the events with respect to the patch size results in information loss. In particular, the spatial origin of each local cloud in relation to the absolute (x,y) pixel grid is disregarded. Thus, the LERT module adds a positional embedding (\emph{i.e.} a learnt linear embedding) to each Patch Feature, incorporating its spatial location within the predefined 2D grid. Positional embeddings supply the model with spatial distribution information about the tokens, they are learnt during training, and they are unique for every patch coordinate. 
A detailed schematic of the LERT embedding can be seen in Figure \ref{fig:embedding_strategy}.

\subsection{ALERT: Asynchronous Token Update}
\subsubsection{Introduction to the ALERT module}
The LERT module is built for synchronous processing.
As such, it does not take full advantage of the event-based nature of the data: it does not naturally update tokens when new events show up.
Instead, it starts from scratch and recalculates all tokens for every new piece of data, which means it must collect events before it can process them.

We thus introduce the Asynchronous LERT (ALERT) Module, which is used during inference.
This variation is designed to efficiently update the token sequence as new events occur (asynchronous processing), making it particularly suited for online inference of event-based data.
To achieve this, the ALERT module requires three key modifications during training and/or inference: handling continuous values of time, updating tokens with each new event, and memorizing past information.

\subsubsection{Handling Continuous Time Values}
The LERT module alters the time coordinates of events in a sequence, shifting each input stream so that the first event starts at time $0$.
This approach introduces inefficiencies in real-time asynchronous processing, as it hinders the feasibility of recurrent or iterative processing of events.
Our proposed solution consists in encoding the time value by using a periodic function.
We achieve this by representing time with two sinusoidal waves: \( t_x(t) = \alpha \cos(2\pi ft + \phi) \), and \( t_y(t) = \alpha \sin(2\pi ft + \phi) \); taking inspiration from the positional encoding originally used in Transformers.
The amplitude ($\alpha$), frequency ($f$) and phase ($\phi$) of these waves are constant and tuned for each application.
The \emph{Time Encoded} (\textbf{TE}) LERT (TELERT) Module applies this periodical wrap prior to the PointNet model. So, TELERT must be the module used during training if ALERT is used for inference.
The only change made before the channel-wise maximum operation between the LERT and TELERT/ALERT modules is this time encoding (see Figure \ref{fig:featuregen} b). 
Next, we present a method applied at token stage that emulates a sliding buffer in input during prediction.

\subsubsection{Asynchronous Event-Based Token Update}
The ALERT module is used for inference, and continuously updates the tokens as new events arrive, whenever $k$ ($k \in \mathbb{Z}^{+}; k >= 1$) events have been triggered.
When decomposing it, a token update requires both (1) adding information from new event(s) and (2) forgetting information from the oldest events.

\textbf{Add Information}: The $k$ new event(s) are processed by TELERT. The resulting patch token is the channel-wise maximum between the previous patch token and the $k$ new event feature vector(s).

\textbf{Forgetting information using Old Maximum Value Decay (OMVD)}: The ``age" of the last update from each tokens' channel is simply tracked by a counter.
All channels whose associated counter channel is higher than a certain threshold undergo decay.
All tokens are subject to this decay, not only the newly updated one, as the oldest information might be stored in different embeddings.
The decay is applied in the form of an exponential ($\times \text{e}^{-\lambda}$), where the decay rate $\lambda$ can be tuned for each model or scene characteristics, and could potentially be learned; the pseudocode can be found in Appendix \ref{sec:appendix:alert}.
This mechanism draws inspiration from the ``leakage" concept in SNNs literature \cite{bouvier_spiking_2019}, but instead of applying it to all neurons/features of the model, we apply it only on tokens, to minimize compute.

\subsubsection{Token Memorization and Activity Status}
In ALERT, each new event triggers a corresponding patch token update.
No raw event needs to be buffered.
Instead, all tokens and associated age counter values are memorized throughout the entire runtime of the model, their patch being active or not.
The non-activeness of the tokens is tracked using an event count for each patch.
This counter is updated ($+1$ or $-1$) every time the token is updated (added information or OMVD, respectively). 

\subsection{(A)LERT-Transformer}
\subsubsection{Transformer Encoder for Classification}
To evaluate the proposed LERT and ALERT modules, we plug them to a Vision Transformer model trained from scratch for classification tasks.
As our focus is to assess the functionality of the proposed learnt synchronous and asynchronous embedding modules, we deploy a stack of standard transformer encoders, as in the original ViT \cite{vit}.
The final classification head simply consists of a single linear layer followed by a SoftMax activation function.

\subsubsection{Synchronous Training and Asynchronous Inference}
During training and for synchronous inference (using LERT or TELERT), all samples are processed independently from each other.
So, classic GPU/TPU accelerated processing is done as usual.
Once TELERT is trained, the embedding module can be converted to ALERT for inference by adding the event update functionalities (adding information and OMVD).

The ALERT version, used only for inference, can process events continuously.
However, applying the feature generator on every single event without distributed processing is extremely time consuming.
Hence, for simulation purposes, the asynchronous inference pipeline is assessed by updating and processing the tokens on-demand every $\Delta t$.
This has no impact on the resulting accuracy, as the mathematics of the asynchronous update are ultimately not altered.
Proof can be found in Appendix \ref{appendix:sec:maxproof}.

%% file: sections/results.tex
%
%

\section{Experiments}
\label{experiments}

\begin{table*}[t]
    \vskip -0.1in
    \caption{Classification performance and complexity on the DVS128Gesture dataset. Event representation meanings: (F=Frames, T=Tokens, E=Events). \emph{Online} refers to the capacity of achieving the said accuracy in a continuous inference paradigm. \emph{Aync.} refers to the event-driven processing nature of the network, where each event can be processed individually. TtA means \emph{Time to Accuracy}, and it is the combination of \emph{input accumulation time} ($\mathbf{t_{in}}$) and \emph{inference time} ($\mathbf{t_p})$.}
    \centering
    \vskip 0.1in
    \resizebox{0.95\textwidth}{!}{
    \begin{tabular}{c|c|c|c|c|c}

        \textbf{Model}                                      & \textbf{Event Repr.}               & \textbf{Async.}                          & \textbf{Online}                            & \textbf{TtA}  & \textbf{Accuracy}                                                                                                                                                                                                                                                                                                      \\ \midrule \midrule
        3D-CNN \cite{innocenti_CNNframesGesture}            & F                                  & \textcolor{red}{\xmark}                  & \textcolor{red}{\xmark}                    & File          & $99.6$\% (FVA)                                                                                                                                                                                                                                                                                                         \\ \midrule

        GET \cite{GET_peng}                                 & F $\rightarrow$ T                  & \textcolor{red}{\xmark}                  & \textcolor{red}{\xmark}                    & File          & $97.9$\% (FVA)                                                                                                                                                                                                                                                                                                         \\ \midrule
        PointNet++ \cite{wang_clouds}                       & E                                  & \textcolor{red}{\xmark}                  & \textcolor{green}{\cmark}                  & $143$ms       & $95.3$\% (NVA)                                                                                                                                                                                                                                                                                                         \\ \midrule

        EventTransAct \cite{Blegiers2023EventTransAct}                       & F $\rightarrow$ T & \textcolor{red}{\xmark}                  & \textcolor{green}{\cmark}                  & $400$ms+$t_p$ & $97.9$\% (NVA) \\ \midrule 
        
        \multirow{2}{*}{Event Tr. \cite{Sabater_2022_CVPR}} & \multirow{2}{*}{F $\rightarrow$ T} & \multirow{2}{*}{\textcolor{red}{\xmark}} & \multirow{2}{*}{\textcolor{green}{\cmark}} & $480$ms+$t_p$ & $94.4$\% (NVA)\tablefootnote{For a fair comparison, we report the accuracy when evaluating their model on full files of DVSGesture test set. It leads to a $1.8$\% difference when compared with the original accuracy presented in \cite{Sabater_2022_CVPR}, where only the central $480$ms of every file were used.} \\ \cline{5-6} 

                                                            &                                    &                                          &                                            & $24$ms+$t_p$  & $79.8$\% (SA)\tablefootnote{\label{note3}The reported SA is obtained when the model makes a prediciton every 24ms, and latent vectors are reset every 20 samples.}                                                                                                                                                       \\ \midrule
        CNN + LSTM \cite{innocenti_CNNframesGesture}        & F                                  & \textcolor{red}{\xmark}                  & \textcolor{green}{\cmark}                  & $500$ms+$t_p$ & $97.7$\% (SA)                                                                                                                                                                                                                                                                                                          \\  \midrule \midrule
        \multirow{2}{*}{LERT-Tr. | RM} & E $\rightarrow$ T                  & \textcolor{red}{\xmark}                  & \textcolor{red}{\xmark}                    & File          & $96.2$\% (FVA)                                                                                                                                                                                                                                                                                                         \\ \cline{2-6}
        & E $\rightarrow$ T   & \textcolor{red}{\xmark}   & \textcolor{green}{\cmark}                                                                                                                & $132$ms+$t_p$                                                              & $88.6$\% (SA)                                                                                                                                                                                                                                                                                                                             \\ \midrule
        \multirow{2}{*}{ALERT-Tr. | RM} & E $\rightarrow$ T                  & \textcolor{green}{\cmark}                & \textcolor{red}{\xmark}                    & File          & $94.1$\% (FVA)                                                                                                                                                                                                                                                                                                         \\ \cline{2-6}
        & E $\rightarrow$ T                  & \textcolor{green}{\cmark}                & \textcolor{green}{\cmark}                  & $9.6$ms ($t_p$)      & $84.6$\% (SA)                                                                                                                                                                                                                                                                                                          \\ \midrule \midrule
        \multirow{2}{*}{LERT-Tr. | LMM} & E $\rightarrow$ T                  & \textcolor{red}{\xmark}                  & \textcolor{red}{\xmark}                    & File          & $92.0$\% (FVA)                                                                                                                                                                                                                                                                                                         \\ \cline{2-6}
        & E $\rightarrow$ T   & \textcolor{red}{\xmark}   & \textcolor{green}{\cmark}                                                                                                                & $132$ms+$t_p$                                                              & $83.1$\% (SA)                                                                                                                                                                                                                                                                                                                             \\ \midrule
        \multirow{2}{*}{ALERT-Tr. | LMM} & E $\rightarrow$ T                  & \textcolor{green}{\cmark}                & \textcolor{red}{\xmark}                    & File          & $89.2$\% (FVA)                                                                                                                                                                                                                                                                                                         \\ \cline{2-6}
        & E $\rightarrow$ T                  & \textcolor{green}{\cmark}                & \textcolor{green}{\cmark}                  & $6.0$ms  ($t_p$)     & $72.9$\% (SA)                                                                                                                                                                                                                                                                                                          \\ \midrule \midrule
    \end{tabular}
    }
    \label{tab:dvsSOTA}
    \vskip -0.1in
\end{table*}

\subsection{Experimental Setup}
\subsubsection{Datasets}
We validate our approach through two classification tasks: action recognition and binary classification.
For action recognition, we utilize the 11-class DVS128Gesture dataset \cite{dvsgesture}, consisting of 2 to 6-seconds recordings from $29$ users.
For training (TE)LERT-Transformers, our input samples consist in Constant Count (CCIM) event streams of $Ne=8192$ events each, with the events being randomly sampled in sequence from the original recordings.
A file usually contains far more than $8192$ events, thus ensuring a huge amount of possible random samples of $Ne$ consecutive events.
For binary classification, the Prophesee's N-Cars dataset \cite{HATS_sironi} is used.
Each file is a $100$ms recording containing either a car or a scene without a car (``background").
In this case our input training samples correspond to one recording, using Constant Time (CTIM) event streams with variable number of events.
We use the original train and test splits for both datasets, and during asynchronous inference simulation with ALERT, we treat the entire test files as continuous sequences of events.

\subsubsection{Performance Metrics}
We describe here the metrics used to evaluate our hybrid asynchronous to synchronous network.

\textbf{Accuracy.} It is common practice in the literature to process several samples sequentially before settling on a decision (a class) \cite{Sabater_2022_CVPR, wang_clouds} -- which we denote \emph{N-Voting Accuracy} (\textbf{NVA}) --, or to estimate the class over full files \cite{GET_peng, innocenti_CNNframesGesture}.
Because our model does not rely on any recurrence between two consecutive input samples, it can process each sample individually and independently.
So, to compare with these works, we evaluate our model by voting over all predictions for each file, denoted \emph{File Voting Accuracy} (\textbf{FVA}).
However, to achieve ultra-low latency, providing a decision with every single sample is crucial.
Hence, we also show the average accuracy over all samples analyzed independently, which we denote as \emph{Sample Accuracy} (\textbf{SA}).
Using SA results in lower accuracy, but enables ultra-low latency.

\textbf{Time to Accuracy (TtA).} The TtA represents the minimum amount of time needed  to classify.
This value depends on both the accuracy measurement method and the \emph{total latency} ($\mathbf{t_{lat}}$) of the network (which equals \emph{input accumulation time} ($\mathbf{t_{in}}$) plus \emph{inference time} ($\mathbf{t_p})$).
In FVA, one needs to get predictions over the full file to reach a conclusion.
So, the TtA is the total duration of a file.
In an online setup (for real-world operation), the accuracy can be evaluated with different strategies, for instance using a sliding buffer for voting over N previous predictions \cite{wang_clouds} (NVA).
We target flexible and ultra-low latency and therefore consider the prediction of individual samples. So, we use the SA method where TtA equals $t_{lat}$.

\textbf{Complexity.}
A good proxy for the time complexity is the number of operations per sample (FLOPs), which is indenpent of the employed hardware.
We also report $t_p$ for our model on an NVIDIA RTX 3080.
The space complexity is associated to the number of parameters of a model.

\subsubsection{Implementation Details}
We present two models: a high-performing \emph{Reference Model} (\textbf{RM}) and a \emph{Low-Memory Model} (\textbf{LMM}) which is a reduced version of the same model.
The RM consists in a 5-layer LERT module and a 4-layer Transformer Encoder with $8$ attention heads and a token width of $512$.
The LMM features a small 2-layer LERT module and a (2 layers, 4 heads, 128-token width) Transformer Encoder.
All models are implemented using \texttt{PyTorch} version 2.1. \cite{paszke_pytorch_2019}.
We train all networks with the cross-entropy loss and the LAMB optimizer \cite{you_large_2020}.

\subsection{Experimental Results}
\subsubsection{LERT-Transformer: Synchronous}
Our RM model showcases noteworthy performance, with an FVA of $96.2$\% on the DVSGesture dataset (see Table \ref{tab:dvsSOTA}).
When measured with SA, the accuracy drops to 88.6\%, but with a much lower latency of $141.6$ms ($t_{in}$=$132$ms$\ +\ t_p$=$9.6$ms) on average.
The RM is relatively complex, running at $1.299$MFLOPs per event, and a total of $13.96$M parameters for the full (A)LERT-Transformer model (see Table \ref{tab:complex-dvs}). The LERT module individually uses $1.218$MFLOPs/event and $1.41$M parameters.
The LMM reduces this complexity by $24$x to an impressive $0.566$M parameters, with the LERT module requiring $264$x less operation per event ($4.0$kFLOPs per event), and $0.04$M parameters. The module together with the transformer runs at $7.4$kFLOPs per event.
LMM still shows a competitive FVA of $92.0$\% on the action recognition task.
On the binary classification task with the N-Cars dataset, LMM achieves an accuracy of $85.6$\% with $0.54$M parameters and an average of only $34.28$M total FLOPs per $100ms$. Note that the average FLOP count per sample varies across different datasets, as the number of events per sample differs depending on sensor resolution and scene dynamics. On the contrary, the FLOPs for processing a single event with our model always remain the same, independently of the dataset.

\begin{table}[h]
    \vskip -0.2in
    \caption{Model complexity on the DVS128Gesture dataset. Here, LERT (RM/LMM) refers only to the LERT module part of the associated model, while RM and LMM refer to the entire model (LERT-Transformer). $t_p$ refers to the inference time, \emph{i.e.} the time ALERT-Transformer takes to process a sample of $8192$ events.}
    \centering
    \vskip 0.1in
    \resizebox{0.95\columnwidth}{!}{
    \begin{tabular}{c|c|c|c|c}

        \multirow{2}{*}{\textbf{Model}} & \multirow{2}{*}{\textbf{\#Param}} & \multicolumn{2}{c|}{\textbf{FLOPs per}} & \multirow{2}{*}{\textbf{$t_p$}}                   \\ \cline{3-4}
                                        &                                   & event & sample  &  \\ \midrule \midrule
        LERT (RM)  & $1.41$M & $1.218$M & $8.83$G & $5.8$ms  \\ \midrule
        RM         & $13.96$M & $1.299$M & $9.42$G & $9.6$ms   \\ \midrule
        LERT (LMM) & $0.04$M & $0.0040$M & $0.03$G & $3.9$ms  \\ \midrule
        LMM        & $0.57$M & $0.0074$M & $0.06$G & $6.0$ms  \\  \midrule \midrule
    \end{tabular}
    }
    \label{tab:complex-dvs}
\end{table}

\subsubsection{ALERT-Transformer: Asynchronous}
Moving from the synchronous LERT to the TELERT module, we noticed a negligible relative accuracy drop in FVA of 0.4\% and 0\% for the RM ($95.8\%$) and LMM ($92.0\%$) respectively (not shown in Table \ref{tab:dvsSOTA}).
This illustrates that time wrapping with sinusoidal encoding is an easy but efficient solution for bounded representation of time values.
From synchronous LERT to ALERT inferences (adding TE and decay) the FVA accuracy relatively decreases by 2\% and 3\% for the RM ($94.1\%$) and LMM ($89.2\%$) versions, respectively.
Nevertheless, this change in the model now allows for \textbf{asynchronous updating of the tokens}.
The latency is entirely customizable as the tokens are continuously updated and can be processed on demand (the asynchronous to synchronous concept illustrated in Figure \ref{fig:async_to_syn}).
The inference times $t_{p}$ for processing $Ne=8192$ events are $9.6$ms and $6.0$ms for the full RM and LMM, respectively.
In the RM (LMM), the (A)LERT module takes $5.8$ms ($3.9$ms) against the $3.8$ms ($2.1$ms) for Transformer and Head.
Simulating this asynchronous low-latency scenario, the RM and LMM models achieve SA accuracies of $84.6\%$ and $72.9\%$, respectively.
Because the (A)LERT module can be run event by event, we argue that using an event-driven asynchronous sparse AI accelerator could reduce the (A)LERT inference time to a few microseconds, thus enabling an on-demand ultra-low latency of $3.8$ms ($2.1$ms) for the RM (LMM).

\subsubsection{Comparison with State-of-the-Art}

\begin{table}[t]
    \vskip -0.1in
    \caption{Classification performance and latency on the N-Cars dataset. Comparison with state-of-the-art asynchronous models.}
    \centering
    \vskip 0.1in
    \resizebox{0.95\columnwidth}{!}{
    \begin{tabular}{c|c|c}

        \textbf{Model}                      & \textbf{Acc.}                               & \textbf{MFLOPs/ev}      \\ \midrule \midrule
        HOTS \cite{hots}                    & $62.4$\%                                    & $14.0$                  \\ \midrule
        HATS \cite{HATS_sironi}             & $90.2$\%                                    & $0.03$                  \\ \midrule
        Async. Sparse CNNs                  & \multirow{2}{*}{$\mathbf{94.4}$\textbf{\%}} & \multirow{2}{*}{$21.5$} \\
        \cite{messikommer}                  &                                             &                         \\ \midrule
        YOLE \cite{cannici2019asynchronous} & $92.7$\%                                    & $328.2$                 \\ \midrule
        EST \cite{gehrig_end--end_2019}     & $92.5$\%                                    & $1050$                  \\ \midrule
        ALERT-Tr. | LMM (Ours)              & $85.6$\%                                    & $\mathbf{0.0074}$       \\ \midrule \midrule
    \end{tabular}
    }
    \label{tab:CarsSOTA}
    \vskip -0.2in
\end{table}

Table \ref{tab:dvsSOTA} presents a comparative analysis with state-of-the-art for action recognition models.
Many competitors lack the ability to process events asynchronously and cannot operate continuously (Online).
PointNet++ \cite{wang_clouds} still exhibits a low latency of $143$ms for an NVA accuracy of $95.3$\% thanks to their sliding buffer voting and averaging strategy for improving accuracy.
Note that our ALERT module could be deployed in their model, which would enable asynchronous PointNet++ processing instead of the current update every $25$ms.
EventNet \cite{sekikawa_eventnet_2019}, a comparable model with asynchronous to synchronous capability, has limited performance on complex tasks as it applies a single PointNet for the full input event stream spatial dimension.
Accuracy measures on gesture recognition has thus not been reported by the authors.
This was one of our motivations towards splitting the input window along the spatial dimension.
EventTransAct \cite{Blegiers2023EventTransAct} processes individual clips containing 16 frames of $5$ms. Every 5 clips, a prediction is obtained through mean aggregation, leading to a latency of $400$ms, and a NVA of $97.9$\%. 
\cite{Sabater_2022_CVPR} propose a recurrent Transformer model that integrates events into frames and processes them every $24$ms.
However, due to reccurence, predictions are originally obtained every 20 samples ($t_{in}=480$ms), introducing significant latency.
When using their model by requesting a prediction every $24$ms (keeping the 20-sample recurrency) and applying SA the accuracy drops to $79.8$\%\footnotemark[3].
In this situation, our ALERT-Transformer RM outperforms theirs with an accuracy of $84.6$\%, but at the price of $1.299$MFLOPs/event.
Even though our LMM requires only $0.0074$MFLOPs/event, in this case, it struggles with a $72.9$\% SA.
But the different hyperparameter searches realized during our study revealed a major cost/accuracy tradeoff (see Appendix \ref{appendix:sec:ablation:hyperparams}).

Furthermore, when compared to state-of-the-art asynchronous models on the N-Cars binary classification task, our LMM showcases the lowest $0.0074$MFLOPs/event complexity while achieving a competitive accuracy (see Table \ref{tab:CarsSOTA}).
This represents a $75$\% decrease in complexity compared to the previous state-of-the-art \cite{HATS_sironi}, with only a $5.11$\% decrease in accuracy.

Overall, the proposed ALERT is a compelling event embedding module for Transformer models.
It inherits event-based data properties, allows for highly flexible latency on-demand compute, and enables friction-less end-to-end learning.

\subsection{Ablation Studies}
We present the effects several hyperparameters have on the RM accuracy and complexity. Accuracies are depicted on the DVS Gesture dataset.
We refer the reader to Appendix \ref{appendix:sec:ablation:visualizations} for a better visualization of the impact these parameters have on the patch sequences and the model.

\begin{figure}[h]
    \begin{center}
        \centerline{\includegraphics[width=\columnwidth]{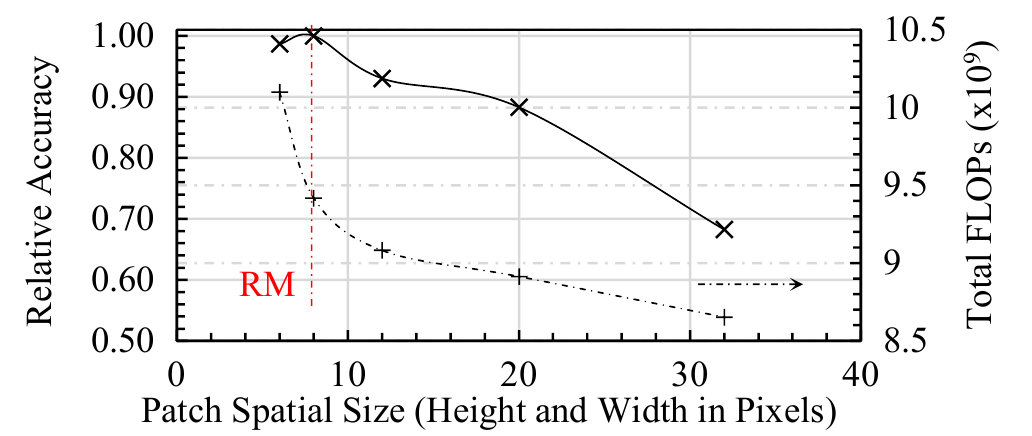}}
        \caption{Influence of the patch size on accuracy ($\times$) and complexity ($+$) (GFLOPs).
        Input resolution is $128\times128$ pixels.
        The RM is set as the baseline, hence has a 1.0 relative accuracy.
        }
        \label{fig:patch_size_variation_graph}
    \end{center}
    \vskip -0.2in
\end{figure}

\textbf{Patch Size.} The ALERT module spatially splits the input event stream in several smaller point clouds.
Figure \ref{fig:patch_size_variation_graph} depicts the variation of accuracy and total FLOPs for RM with respect to the patch size (height and width, in pixels).
Smaller patches yield better models, but they are more complex and, as such, slower.
The low accuracy of models with bigger patch size is predictable, as attention networks are known to perform worse with too short sequences.
As the patch size increases, the number of available sub clouds -- and consequential tokens -- decreases, causing the poor performance of the Transformer classification network.
Therefore, a tradeoff is necessary to find the optimum for each application and sensor resolutions.

\begin{figure}[h]
    \begin{center}
        \centerline{\includegraphics[width=\columnwidth]{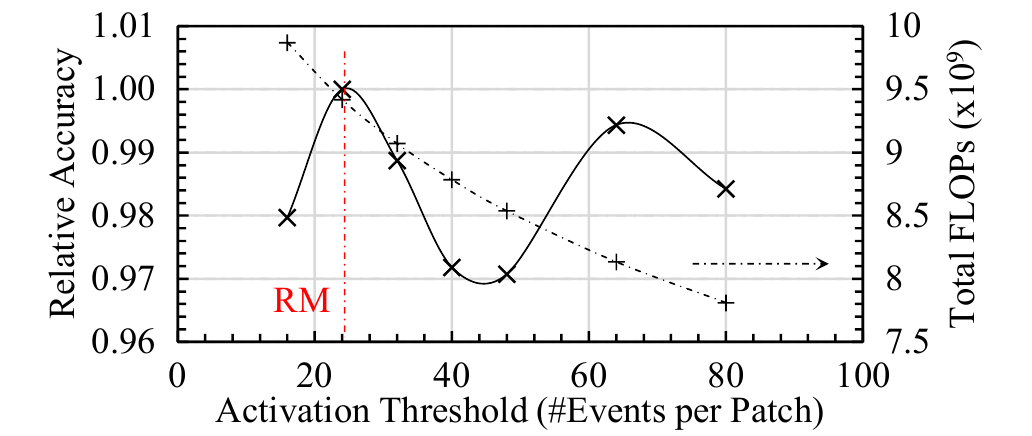}}
        \caption{Influence of the patch activation threshold size on accuracy and complexity (GFLOPs). The RM is set as the baseline, hence has a 1.0 relative accuracy.}
        \label{fig:activation_rate_variation_graph}
    \end{center}
    \vskip -0.2in
\end{figure}

\textbf{Active Patch Threshold.} Understanding the role of the activation threshold in the (A)LERT-Transformer is key for using it efficiently.
This threshold -- defined in number of events per patch -- determines the portion of patches considered as active.
Interestingly, according to Figure \ref{fig:activation_rate_variation_graph} comparable accuracies are achieved with both low and high thresholds.
Nevertheless, higher thresholds result in less complex models because the FLOPs per processed event remains constant, while less events are processed.
In RM, the Transformer model is so small that the embedding part represents $94$\% of the total FLOPs of the model for $Ne=8192$ events.
Therefore, choosing a high threshold may be an important decision when a low complexity model with high accuracy is required.

\begin{figure}[b]
    \begin{center}
        \centerline{\includegraphics[width=\columnwidth]{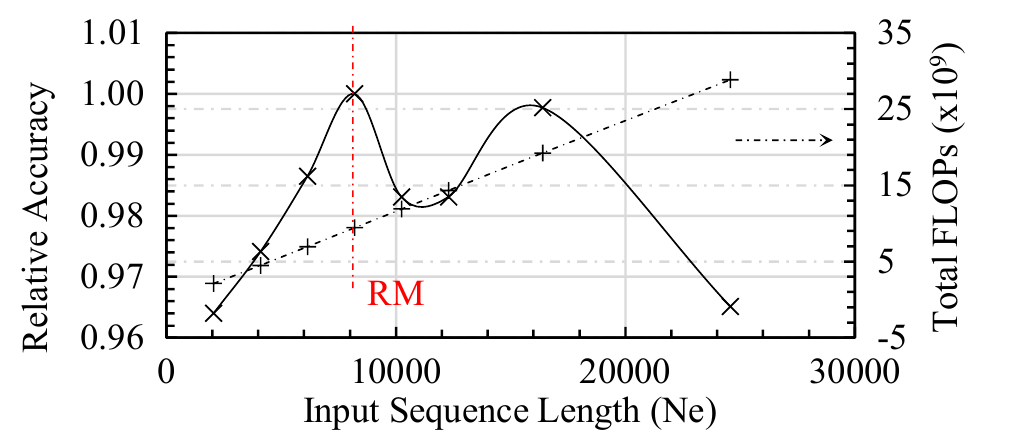}}
        \caption{Influence of the input sequence length size on accuracy and complexity (GFLOPs). The RM is set as the baseline, hence has a 1.0 relative accuracy.}
        \label{fig:sequence_length_variation_graph}
    \end{center}
    \vskip -0.2in
\end{figure}

\textbf{Input Sequence Length.} The (TE)LERT module can be trained and evaluated in Constant Count or Constant Time input modes, and is thus suited for processing input sequences of varying lengths ($Ne$ or $\Delta_t$, respectively).
In Figure \ref{fig:sequence_length_variation_graph}a, we show how $Ne$ impacts the CCIM model's performance.
There seems to be an optimum point in between $Ne=8$k and $Ne=16$k events for the accuracy. 
However, the total complexity scales linearly with the number of processed events. Hence, our experiments were conducted using $Ne=8192$, which yields $96.2\%$ FVA.
Similarly, for CTIM performance and complexity vary depending on $\Delta_t$ values (see Appendix \ref{sec:appendix:CTIM}). The optimal point in this case is $\Delta_t = 180$ms, where the model reaches $88.5\%$ SA and $96.5\%$ FVA. 
The average time span of a DVS Gesture input sample in CCIM with $Ne = 8192$ fixed events, is of ~$130$ms. It is worth noting that using $\Delta_t=130$ms in CTIM reaches $96.2\%$ FVA, proving the equivalence between both input modes. 
Consequently, the input mode and its length should be adjusted based on task-specific requirements.

\begin{figure}[h!]
    \begin{center}
        \centerline{\includegraphics[width=\columnwidth]{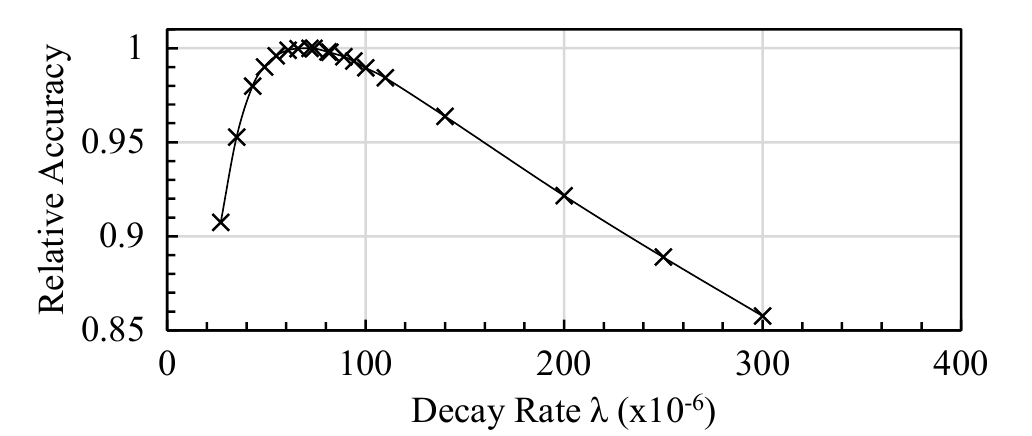}}
        \caption{Decay rate influence on the ALERT accuracy during inference.}
        \label{fig:decayrate}
    \end{center}
    \vskip -0.2in
\end{figure}

\textbf{Old Maximum Value Decay.} The decay rate is applied in the form of an exponential factor whenever a token channel has not been updated for a certain time. An optimal value exists for this decay rate, which should be tuned to simulate the behavior of the synchronous processing paradigm as truthfully as possible. Figure \ref{fig:decayrate} shows the relationship between accuracy and decay rate for RM. Below the optimal rate, our model forgets past information too quickly; above the optimal value too much past information is kept, bringing confusion to the prediction. In the limit where information is not forgotten, \emph{i.e.} $\lambda = 0$, the model reaches a sample accuracy of $25.9\%$, proving OMVD is a key component of the ALERT-Transformer.

\textbf{Positional embeddings.} The encodings of the patch coordinates, provide information about the patch tokens spatial distribution in space. Training the LERT-Transformer Reference Model (RM) on the DVSGesture dataset without these positional encodings, leads to accuracies of $80.0\%$ SA and $95.1\%$ FVA (versus $88.6\%$ SA, $96.2\%$ FVA with positional embeddings). During inference they can be built as a simple Look-Up table, introducing a negligible overhead to the total cost of the model.

%% file: sections/discussion.tex
%
%

\subsection{Limitations and Potential Applications}
\label{discussion}

While the outcomes of this study showcase the viability of the proposed ALERT module for event data in Transformer-based computer vision pipelines, there are some limitations and room for improvement.
The current validation focuses on a Transformer architecture for classification tasks.
However, the tokens do not have to be used only by Transformers, and our method could be easily integrated with Convolutional Neural Networks (CNNs) or other architectures.
Similarly, we are also working on the applicability of this learnt representation for other tasks than those used for benchmarking, such as generative ones.
We see several directions for exploiting our research: trying to improve accuracy by adopting advanced strategies akin to concurrent processing of multiple samples \cite{GET_peng} or using of memory tokens \cite{Sabater_2022_CVPR}; or exploring PointNet-only architectures based on the ALERT module \cite{wang_clouds, sekikawa_eventnet_2019}.

The demonstrated learnt token sequence in the ALERT-Transformer model holds promise for diverse applications relying on computer vision, especially for edge solutions.
Its adaptability across various pipelines, beyond transformer architectures, suggests utility in multiple tasks where event vision sensors offer advantages over conventional cameras.
The flexibility of the on-demand data processing, and potential resulting energy-efficient capabilities of the ALERT-Transformer create opportunities for real-world deployment in scenarios using advanced systems for computer vision (always-on sensing to wake-up Application Processing Unit).
Future research may explore integrating this hybrid asynchronous to synchronous representation into a broader array of vision tasks, leveraging its strengths for diverse real-time applications, and paving the way for potential hardware implementations.
Also, it is worth noting that our pipeline could be integrated in multi-modal Transformers, seamlessly enabling mixed sensor computer vision.

%% file: sections/conclusions.tex
%
%

\section{Conclusions}
\label{conclusions}
Our research focused on addressing the fundamental challenge of efficiently processing sparse and asynchronous event-based data while leveraging its properties.
Several, simple yet essential, contributions have been introduced: (1) end-to-end trained event-data to feature vectors conversion with the LERT module, and the modified Time Encoded LERT able to deal with continuous time values with negligible decrease of accuracy. (2) Seamless conversion to real-time event-driven processing with Asynchronous LERT, significantly reducing latency and enhancing the model's applicability to real-time applications. (3) An asynchronous sensing to on-demand synchronous processing framework: the ALERT-Transformer. An end-to-end system that ensures continuous and energy-efficient data processing for event-based vision sensors.

The ALERT-Transformer on the gesture recognition task achieves high accuracy ($84.6\%$) with lowest ever latency (less than $9.6$ms) during inference, outperforming comparative models with higher accuracy but slower processing.
The reduced size model performs asynchronous binary classification at the lowest ever cost of $7.4$kFLOPs/ev (to the best of our knowledge).

These contributions allow for significant advancement towards exploiting the potential of sparse heterogeneous multidimensional data, and lay the ground for further advancements in mixed sensors  computer vision.

%% file: sections/appendix.tex
%
%

\section{ALERT Module: Proofs and Definitions}

\subsection{Algorithm: Asynchronous Event-Based Token Updates}
\label{sec:appendix:alert}
The following algorithm presents the mathematical operations performed during a single token update with the ALERT module in inference mode. This update corresponds to a token, $G_a$, from a patch that has a new event, $e_{new}$, whose information needs to be incorporated. The Old Maximum Value Decay strategy is also applied to this token. OMVD consists in applying a decay to old values and, although not shown in the algorithm, it is applied to in all tokens, not only $G_a$. Below, the function $f(e)$ represents the shared MLP that builds a feature for each event; $N$ is the threshold for OMVD to be applied; and $counter_a$ keeps track of the latest updates in all channels of $G_a$.
\begin{algorithm}
    \caption{Asynchronous Event-Based Update of a Patch Token with Old Maximum Value Decay}
    \begin{algorithmic}
        \REQUIRE $e_{new}$, $counter_a$, $G_a$, $N$
        \FOR{$C$ in embbedings dimension}
        \STATE $G_a[C] \gets \text{max}(G_a[C],\ f(e_{new})[C])$
        \IF{$G_a[C]\ ==\ f(e_{new})[C]$}
        \STATE $counter_a[C] \gets 0$
        \ELSE
        \STATE $counter_a[C] \gets counter_a[C] + 1$
        \IF{$counter_a[C] > N$}
        \STATE $G_a[C] \gets G_a[C]\times \exp{(-\lambda)}$
        \ENDIF
        \ENDIF
        \ENDFOR
    \end{algorithmic}
\end{algorithm}

\subsection{Proof: Iterative Updating of Tokens}
\label{appendix:sec:maxproof}
Global features in the proposed model are obtained from the channel-wise maximum values amongst all the event features in a cloud.
Let us demonstrate the equivalence between updating the token channels as each events arrives, and creating the tokens by applying the max function to several available new events at once (TELERT module).

Consider a sequence of values $x_1, x_2, \ldots, x_N$, and let $M$ represent the maximum value in the sequence.
In the LERT token creation paradigm, the maximum value is obtained by applying the max function to all values at once:

\begin{equation}
    M_{\text{all}} = \max(x_1, x_2, \ldots, x_N)
\end{equation}

On the other hand, with the ALERT token creation paradigm, the maximum value is obtained by applying the max function iteratively as new values arrive. Define $M_k$ as the maximum value after considering the first $k$ values:

\begin{equation}
    M_k = \max(x_1, x_2, \ldots, x_k)
\end{equation}

After adding a new value $x_{k+1}$ to the sequence, the updated maximum is:

\begin{equation}
    M_{k+1} = \max(M_k, x_{k+1})
\end{equation}

Now, let us show that $M_{\text{all}} = M_N$, demonstrating the equivalence of the two approaches.

\begin{align*}
    M_{\text{all}} & = \max(x_1, x_2, \ldots, x_N)                                   \\
                   & = \max(\max(x_1, x_2, \ldots, x_{N-1}), x_N)                    \\
                   & = \max(\max((\max(\max(x_1, x_2), x_3), \ldots), x_{N-1}), x_N) \\
                   & = M_N
\end{align*}

We have now shown that applying the max function to all values at once is equivalent to applying it iteratively as new values arrive. Furthermore, we would like to proof that the Old Maximum Value Decay strategy leads to congruent results.

Consider a global feature where a specific channel, $M$, has not been updated for a long time, meaning that the channel is taking the information from an old event. In this case, with non-iterative processing, this channel's value would be updated, as the old event feature would not be included in the $max$ operation anymore.
\begin{equation}
    M = \max(x_0, x_1, ..., x_N)\ \ \rightarrow x_{N+1}\ \text{not used}
\end{equation}

Here, $x_i$ represents the corresponding channel from each event feature. Now, if we want to do iterative updating of the tokens, we cannot recompute the $max$ over all values every time. Thus, we find a decay strategy where old channel values lose their relevance over time, instead of completely removing these values.
\begin{equation}
    M = \max(x_0, x_1, ..., x_N, x_{N+1}\times e^{-\lambda})
\end{equation}
As time continues and new events arrive, the oldest events values lose their relevance exponentially.
\begin{equation}
    X = \max(x_0, x_1, ..., x_N, x_{N+1}\times e^{-\lambda}, x_{N+2}\times e^{-2\lambda}, ..., x_{N+l}\times e^{-l\lambda})
\end{equation}
The decay term ensures that the influence of events decreases exponentially over time. Thus, in the limit where time approaches infinity, channels from old event features will tend to zero, not being able to ``win'' in the $max$ function.

The ALERT module methodology comes from merging the iterative addition of information and this decay strategy. The combination of both approaches slightly alters the results from completely computing tokens from scratch, but the difference is negligible when compared to the gains the ALERT module provides.
\begin{align*}
    X_t = & \max(x_{new}, X_{t-1})                    & \text{if }counter_c <\text{ threshold}  \\
    X_t = & \max(x_{new}, X_{t-1}\times e^{-\lambda}) & \text{if }counter_c >=\text{ threshold}
\end{align*}

\newpage
\section{(A)LERT-Transformer: In-Depth Ablation Study}
\label{appendix:sec:ablation}
The following figures and tables show in a visual and detailed manner the influence several hyperparameters have on the ALERT-Transformer model.

\subsection{Patch Size, Activation Rate and Sequence Length}
\label{appendix:sec:ablation:hyperparams}
\begin{table}[h!]
    \caption{\textbf{ALERT Hyperparameter Analysis.} Influence of several key embedding hyperparameters on the performance and complexity of the LERT-Transformer model trained on the DVS Gesture Dataset for action recognition.}
    \label{tab:hyperparams_analysis}
    \centering
    \begin{tabular}{ccc}
        (a) Sequence Length                                                                 & (b) Voxel's Spatial Size & (c) Voxel's Activation Rate \\
                                                                                            &                          &                             \\

        \begin{tabular}[t]{c|c|c}
                  & \textbf{Relative} & \textbf{Total} \\
                  & \textbf{Accuracy} & \textbf{FLOPs} \\
            \midrule \midrule
            2048  & $0.963$\%         & $2.092$G       \\
            4096  & $0.973$\%         & $4.492$G       \\
            6144  & $0.985$\%         & $6.949$G       \\
            8192  & $1.000$\%         & $9.419$G       \\
            10240 & $0.982$\%         & $11.87$G       \\
            12288 & $0.982$\%         & $14.32$G       \\
            16384 & $0.997$\%         & $19.24$G       \\
            24576 & $0.964$\%         & $28.86$G       \\
        \end{tabular}                                       &

        \begin{tabular}[t]{c|c|c}
                       & \textbf{Relative} & \textbf{Total} \\
                       & \textbf{Accuracy} & \textbf{FLOPs} \\
            \midrule \midrule
            $[6,6]$    & $0.985$\%         & $10.10$G       \\
            $[8,8]$    & $1.000$\%         & $9.419$G       \\
            $[12,12]$  & $0.973$\%         & $9.082$G       \\
            $[20,20]$  & $0.985$\%         & $8.914$G       \\
            $[32, 32]$ & $0.999$\%         & $8.653$G       \\
        \end{tabular} &

        \begin{tabular}[t]{c|c|c}
                 & \textbf{Relative} & \textbf{Total} \\
                 & \textbf{Accuracy} & \textbf{FLOPs} \\
            \midrule \midrule
            0.50 & $0.979$\%         & $9.871$G       \\
            0.75 & $1.000$\%         & $9.419$G       \\
            1.00 & $0.988$\%         & $9.071$G       \\
            1.25 & $0.971$\%         & $8.785$G       \\
            1.50 & $0.970$\%         & $8.538$G       \\
            2.00 & $0.993$\%         & $8.134$G       \\
            2.50 & $0.983$\%         & $7.811$G       \\
        \end{tabular}                                                                                                    \\
    \end{tabular}
    \vskip -0.1in
\end{table}

\subsection{Visualization of the Patch Size and Sequence Length}
\label{appendix:sec:ablation:visualizations}

\begin{figure}[h!]
    \vskip 0.2in
    \centering
    \includegraphics[width=0.6\textwidth]{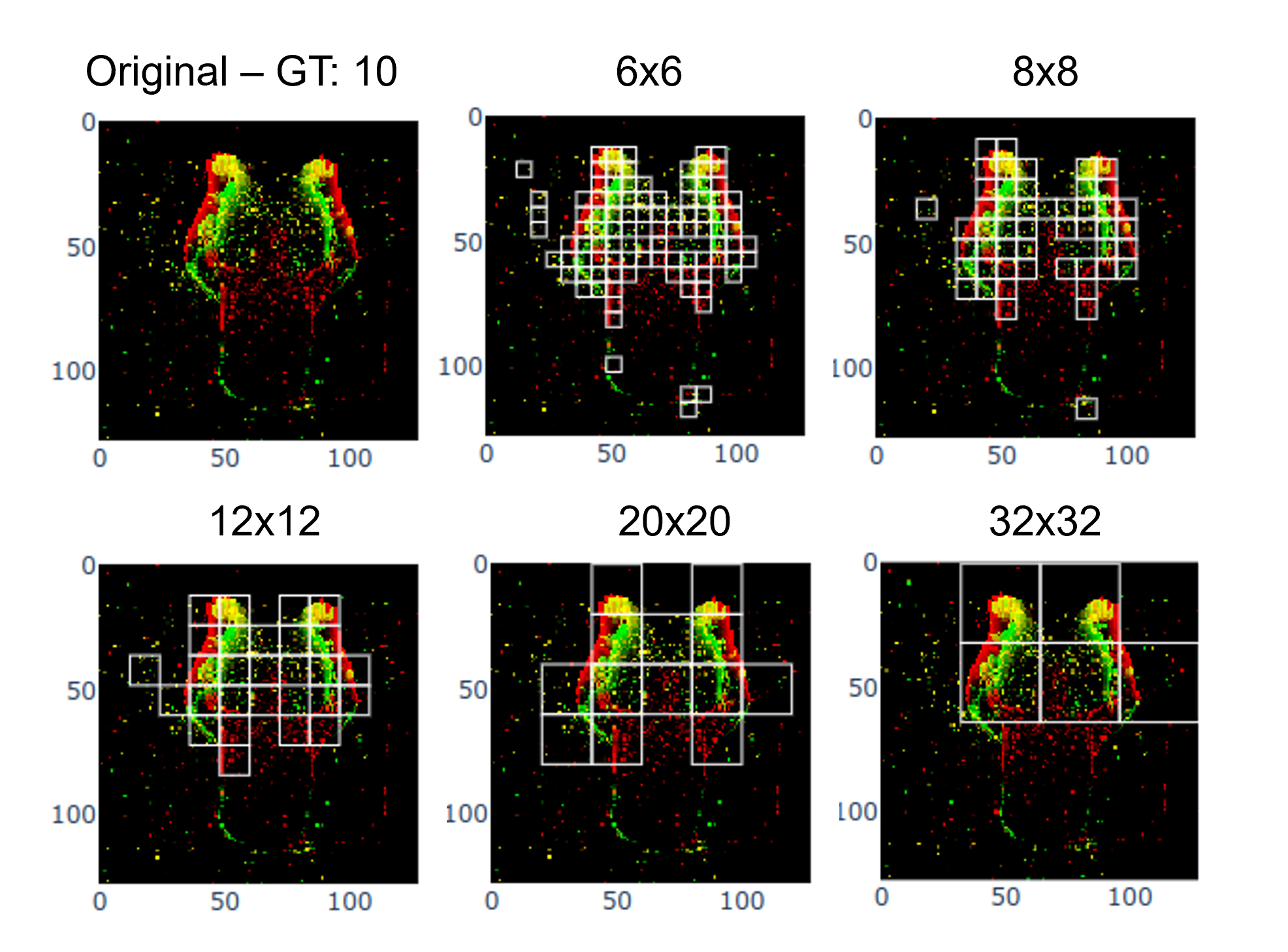}
    \caption{\textbf{Patch Size Visualization:} A simple 2D representation of the same sample with its corresponding active voxels for different patch sizes. This representation allows for a better understanding of the spatial division of the event stream. Each drawn patch corresponds to an active voxel, which events will be processed through the feature generator to obtain a token. The smaller the patch size, the less information a single voxel contains. As a result, with the same processing pipeline, a more detailed representation can be obtained when using smaller patches. The need exists to find a trade-off between the complexity of the model, and the level of detail of the token representations. The sample consists of a stream of $8192$ events from the DVS128 Gesture Dataset.}
    \label{fig:patchsize_vis}
    \vskip -0.2in
\end{figure}

\begin{figure}[h!]
    \vskip 0.2in
    \centering
    \includegraphics[width=0.8\textwidth]{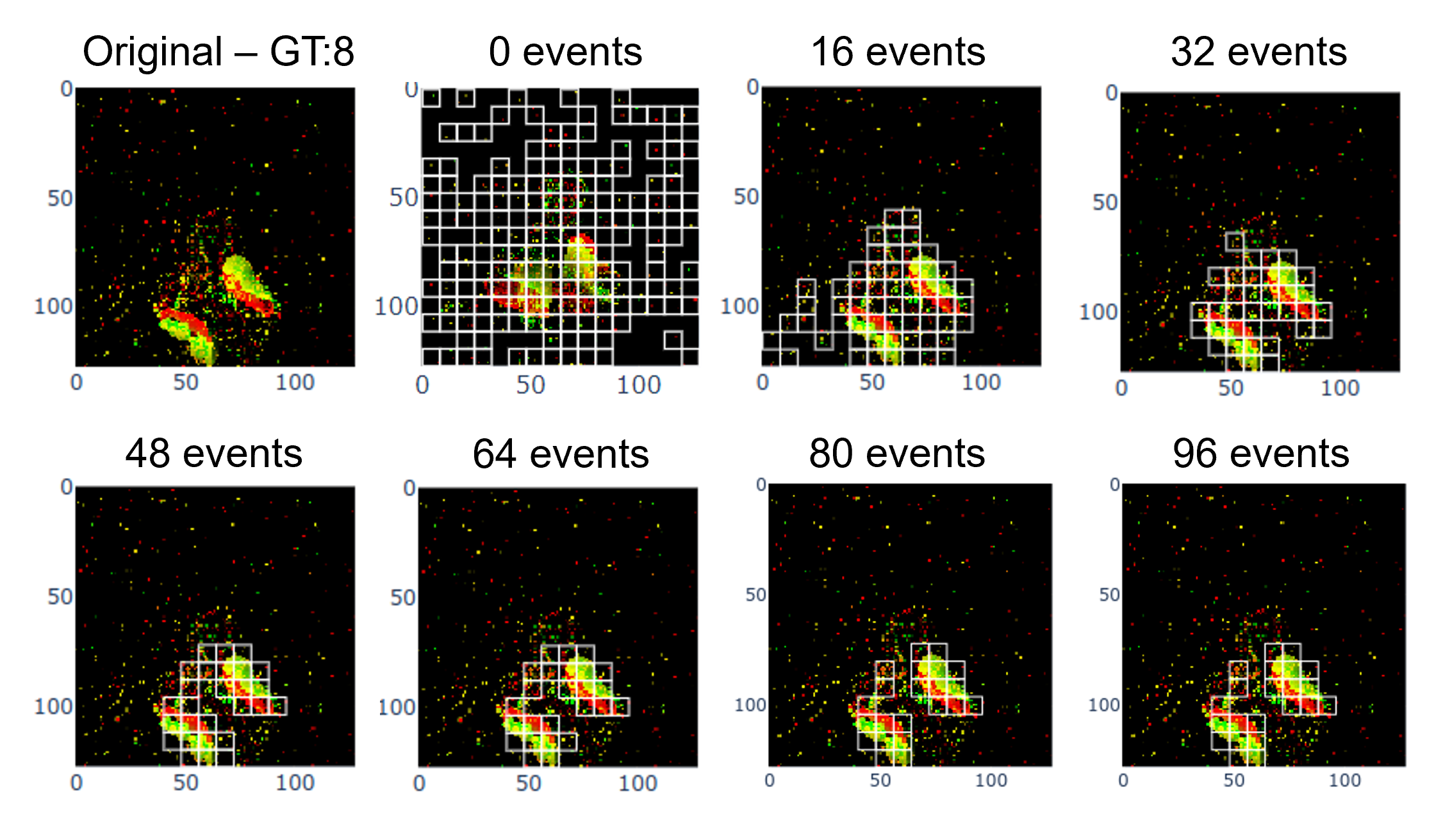}
    \caption{\textbf{Patch Activation Threshold Visualization:} A simple 2D representation of the same sample with its corresponding active patches for the same patch size, but different activation thresholds. This simple representation allows for a straightforward understanding of the necessity of discarding certain patches. Event-based data is inherently sparse, and this should be leveraged when processing it. Eventhough the model performs similarly with different thresholds, it is important to note that processing almost empty patches does not add any relevant information while adding a great computational cost. Discarding sub clouds where no movement occurs removes noise and allows the model to decrease its complexity. The sample consists of a stream of $8192$ events from the DVS128 Gesture Dataset.}
    \label{fig:actrate_vis}
    \vskip -0.2in
\end{figure}

\begin{figure}[h!]
    \vskip 0.2in
    \centering
    \includegraphics[width=0.8\textwidth]{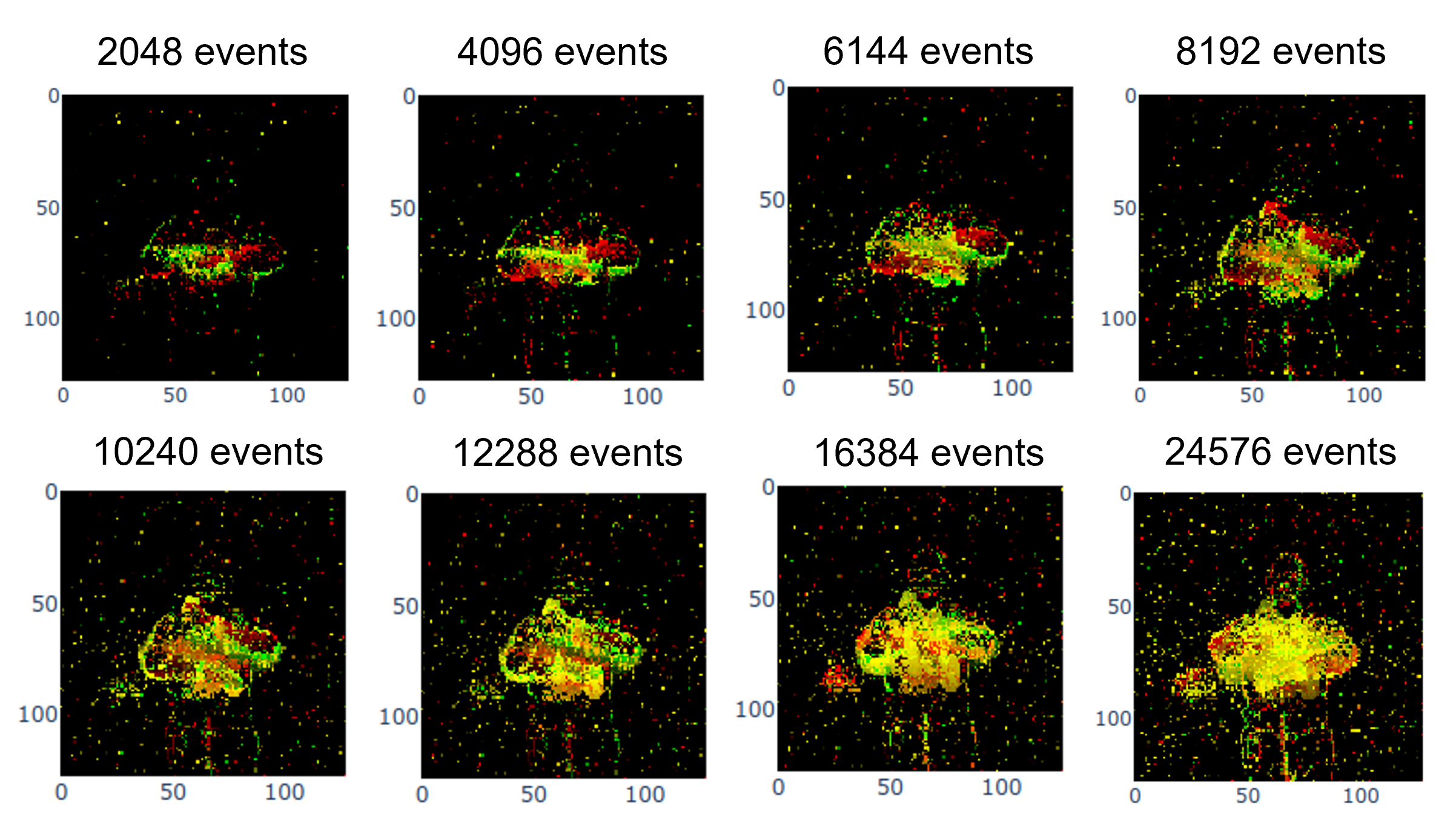}
    \caption{\textbf{Sequence Length Visualization:} A simple 2D representation of samples formed by different number of events. All of the samples share the first event but extend to different lengths. The samples consist of streams of events from the DVS128 Gesture Dataset. Although this frame representation of the samples is \emph{not} used at any point in the proposed pipeline, it serves as visualization of the different sample lengths and the amount of information they carry. This analysis highlights the necessity of selecting input samples appropriately to the dataset and task requirements, as too little information may limit the performance, while too much may be too difficult to learn with a simple model.}
    \label{fig:seqlen}
    \vskip -0.2in
\end{figure}

\newpage
\subsection{Feature Generator Network Depth}
The optimal number of layers for the Feature Generator's MLP strongly depends on other hyperparameters of the model. This value has a direct impact on the model's memory footprint, as it changes the number of parameters of the model. The Feature Generator has two important tunable hyperparameters: the number of layers of the MLP (depth) and the number of neurons in each of these layers. The input and output of the FG have fixed dimensions, being a 4-dimensional point and a 512-dimensional vector, respectively. However, the dimension and number of in-between hidden layers can vary depending on the model's requirements. Three different hyperparameters are used during our implementation to define this configuration:
\begin{itemize}
    \item \emph{Depth:} Number of layers of the MLP. In the implementation, each of these layers comprises a 1D-convolutional layer, batch normalization and an activation function (ReLU).
    \item \emph{Base channels:} Number of neurons of the first hidden layer.
    \item \emph{Expansion Factor ($\beta$):} Determines the increment of neurons from one hidden layer to the next as follows: $hidden_i = \beta \times hidden_{i-1}$.
\end{itemize}

\begin{figure}[h]
    \vskip 0.2in
    \centering
    \includegraphics[width=\textwidth]{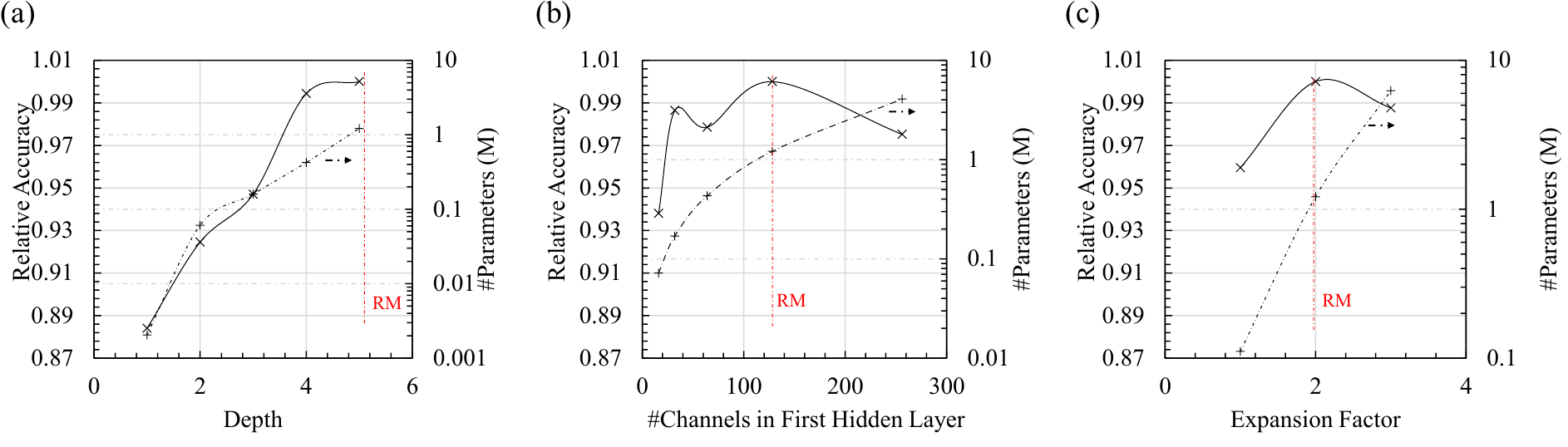}
    \caption[Feature Generator's Depth impact on the Model's Performance]{\textbf{Feature Generator's Depth impact on the Model's Performance.} Overall,the depth of the (A)LERT module has a major influence on the accuracy of the network. The deeper the most accurate. But at the same time, network complexity increases. So a tradeoff needs to be made regarding the depth. On the other hand, the projection dimension does not have the same relationship with accuracy. Indeed, it seems that one should avoid using a too small projection width, but an optimum can be found. The same seems to be true for the expansion factor. The two later points provide an optimism regarding the search for an optimal accuracy to cost ratio at a given depth. Note: the vertical red lines illustrate the RM configuration.}
    \label{fig:hyperparams_MLP}
    \vskip -0.2in
\end{figure}

Figure \ref{fig:hyperparams_MLP} provides an insight on how the modification of these values alters the performance and memory footprint of the model. In general, the fewer layers and number of neurons, the less accurate a model is. However, this decrease in accuracy is quite slow. On the contrary, reducing the number of hidden layers, and the neurons per layer, decreases the size of the model exponentially. The relationship between the metrics and the MLP's hyperparameters highlights the necessity of a trade-off between complexity and accuracy for each model, and the optimal solution will depend on the specific application requirements. Our RM and LMM provide two cases where this trade-off are considered, but further models are to be explored.

\newpage
\subsection{Constant Time Input Mode}
\label{sec:appendix:CTIM}

\begin{figure}[h]
    \vskip 0.2in
    \centering
    \includegraphics[width=0.5\textwidth]{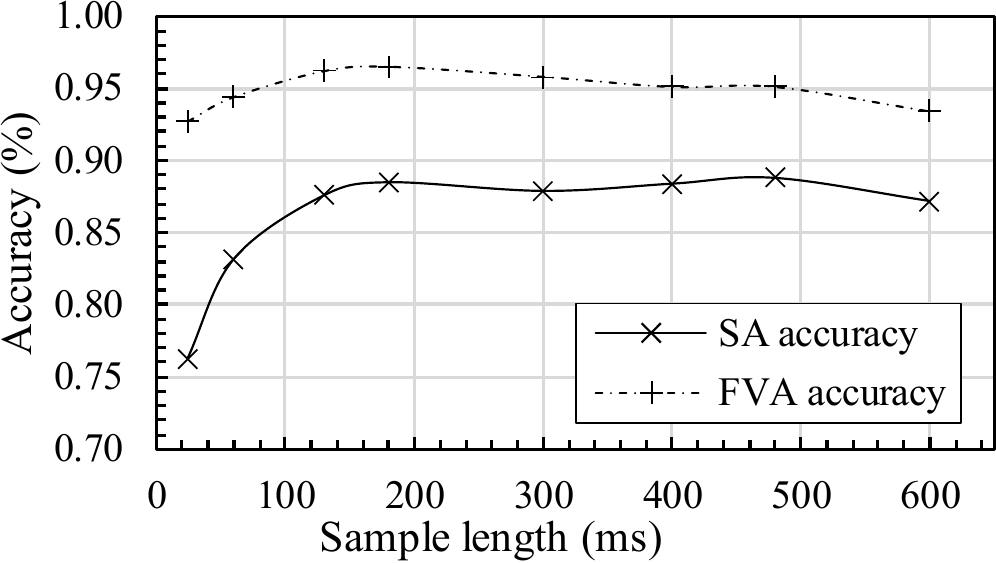}
    \caption[Duration of an Input Sample (in CTIM) impact on the Model's Performance]{\textbf{Duration of an Input Sample (in CTIM) impact on the Model's Performance.}}
    \label{fig:hyperparams_CTIM_duration}
    \vskip -0.2in
\end{figure}